\documentclass[9pt]{article}
\usepackage{spconf,amsmath,graphicx}
\usepackage[numbers]{natbib}
\usepackage{enumitem}
\setlist{nosep, leftmargin=14pt}
\usepackage{url}
\usepackage{subcaption}
\usepackage{amsfonts}
\usepackage{xcolor}
\usepackage{pifont}
\usepackage{mwe} 

\title{AI-Based Thermal Video Analysis in Privacy-Preserving Healthcare: A Case Study on Detecting Time of Birth}
\name{Jorge García-Torres$^{\star}$ \qquad Øyvind Meinich-Bache$^{\star \dagger}$ \qquad Siren Rettedal $^{\ddag \S}$ \qquad Kjersti Engan$^{\star}$}

\address{$^{\star}$ Dept. of Electrical Engineering and Computer Science, University of Stavanger, Norway \\
    $^{\dagger}$ Laerdal Medical AS, Stavanger, Norway \\
    $^{\ddag}$ Dept. for Simulation-based Learning, Stavanger University Hospital, Norway \\
    $^{\S}$ Faculty of Health Sciences, University of Stavanger, Norway
    }
\begin{document}
\ninept
\maketitle
\begin{abstract}
Approximately 10\% of newborns need some assistance to start breathing and 5\% proper ventilation. It is crucial that interventions are initiated as soon as possible after birth. Accurate documentation of Time of Birth (ToB) is thereby essential for documenting and improving newborn resuscitation performance. However, current clinical practices rely on manual recording of ToB, typically with minute precision. In this study, we present an AI-driven, video-based system for automated ToB detection using thermal imaging, designed to preserve the privacy of healthcare providers and mothers by avoiding the use of identifiable visual data. Our approach achieves 91.4\% precision and 97.4\% recall in detecting ToB within thermal video clips during performance evaluation. Additionally, our system successfully identifies ToB in 96\% of test cases with an absolute median deviation of 1 second compared to manual annotations. This method offers a reliable solution for improving ToB documentation and enhancing newborn resuscitation outcomes.
\end{abstract}

\section{Introduction}
\label{sec:intro}

\begin{figure*}[ht]
\centering
\begin{minipage}[b]{.97\linewidth}
  \centering
  \centerline{\includegraphics[width=\linewidth]{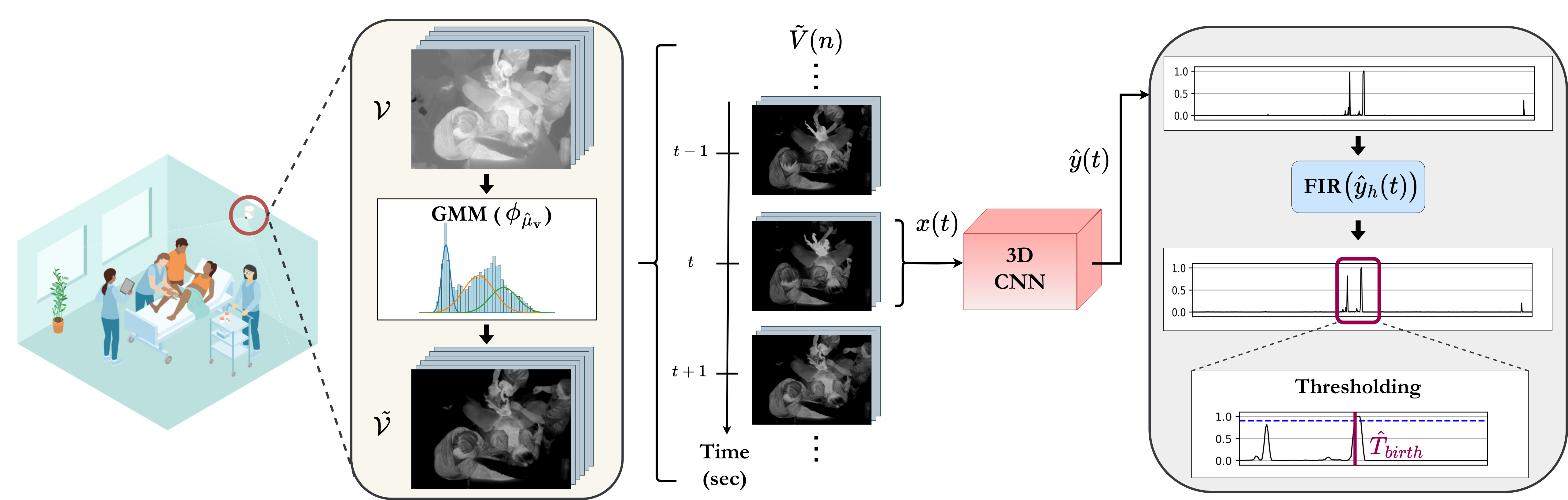}}
\end{minipage}
\caption{Overview of our proposed video-based, AI-driven Time of Birth detector. Birth episodes are recorded by a thermal camera installed on the ceiling. First, thermal videos undergo adaptive normalization using Gaussian Mixture Models, followed by trimming into video clips. At each second, a sliding window containing the previous 25 frames (3 seconds) is utilized as input $x(t)$ to our model for prediction. All the model's output scores $\hat{y}(t)$ are then concatenated and post-processed using a Finite Impulse Response (FIR) filter with a rectangular window. Finally, Time of Birth $\hat{T}_{birth}$ is inferred based on the model's confidence in successfully detecting the birth in the video clips.}
\label{fig:overview}
\end{figure*}

Approximately 10\% of newborns need some assistance to start breathing, and 5\% proper ventilation, with birth asphyxia as one of the leading causes of neonatal mortality, responsible for an estimated 900,000 deaths annually \cite{ersdal2012early, who-asphixia}. 
The Time of Birth (ToB) is defined as the moment when the newborn's head, torso, and nates are fully visible outside the mother's perineum, and mark the critical reference point for resuscitation interventions \cite{branche2020first}. According to neonatal resuscitation guidelines \cite{madar2021european, wyckoff20232022}, Newborn Resuscitation Algorithm Activities (NRAA) should be started within the ``golden minute'' -- the first 60 seconds after birth, as immediate action can significantly reduce the risks of death or long-term complications due to birth asphyxia \cite{ersdal2012early}. The NRAA guidelines are based on best practices and evidence research needs to be sought \cite{McCarthy2013}. NRAA timelines are crucial for retrospective analysis and debriefing, research on optimal birth asphyxia treatment, and potential real-time clinical decision support. Accurate ToB documentation is an essential part of an effective NRAA timeline; however, in clinical practice, ToB is recorded manually, often with minute precision. Such lack of precision reduces the reliability of NRAA timeline documentation.

Artificial Intelligence (AI) has demonstrated promise in automating NRAA timelines \cite{meinich2020activity, garcia-torres2023comparative}, offering a way to streamline the documentation of newborn resuscitation activities. Yet, existing AI-based approaches still depend on manually recorded ToB, limiting the broader applicability of these systems in critical care settings. 

The NewbornTime project \cite{engan2023newborn} seeks to address these gaps by developing a fully automated system that generates accurate NRAA timelines, including ToB and resuscitation activities. One of the key innovations is our privacy-preserving method, which utilizes thermal imaging to detect ToB by identifying the higher skin temperature of newborns compared to other persons in the room. Thermal or infrared (IR) imaging captures the heat emissions, forming images that show temperature differences. This technique ensures the privacy of both healthcare providers and mothers by avoiding identifiable visual data. In healthcare, such technologies can be pivotal in building trust and maintaining compliance with data protection regulations, such as the General Data Protection Regulation (GDPR).

In \cite{garcíatorres2024advancingnewborncareprecise} we demonstrated that AI-driven, image-based systems using individual thermal frames had potential for estimating ToB. However, the lack of temporal context led to imprecision in certain scenarios. To the author's knowledge, no other work on automatic ToB detection exists. The primary contribution of this work is a spatiotemporal AI-based method for ToB detection using privacy-preserving thermal video. By analyzing continuous changes in movement and thermal characteristics, the model captures the dynamic birth process, resulting in a more precise ToB documentation.

\section{Data Material}
\label{sec:dataset}

The dataset contains a total of 321 birth videos captured by thermal cameras. Approximately 80\% of the births occurred in supine position, 15\% in side-lying position, and 5\% in hands-and-knees position. Data were collected in a semiautomatic manner \cite{Brunner2025} at Stavanger University Hospital (SUS), Norway, using a passive thermal module supplied by Mobotix \cite{mobotix:v1.02} and installed in eight delivery rooms. The sensor was mounted to the ceiling, centered above the head of the mother. The recording was triggered when any pixel's temperature exceeded 30$^\circ$C, detecting human presence and streamlining data capture. ToB was manually registered with second precision by midwives or nurse assistants using the Liveborn Observation App \cite{bucher2020digital}, specifically designed to document post-birth events in research projects, providing more accurate ToB logging than standard practice \cite{kolstad2024detection}. Pressing the ``Baby Born'' button in this app automatically saved a 30-minute thermal video generated at 252$\times$336 resolution and 8.33 fps, covering 15 minutes before and after birth. The rest of the video was discarded.

The registration of ToB in Liveborn was sometimes delayed due to all the activities that happened around the birth. Therefore, in this study, manual annotation of the ToB with second precision was done by carefully inspecting the thermal videos.

\section{Methods}
\label{sec:methods}

In this work, we introduce an AI-driven, video-based system that uses spatiotemporal information to estimate ToB. The task is formulated as a binary classification problem where we want to identify the birth within video clips. We propose a pipeline where thermal videos are first normalized with an adaptive method based on Gaussian Mixture Models (GMM) and then streamed into our model using a sliding window with a constant stride for generating predictions. An overview of the proposed method is illustrated in Figure~\ref{fig:overview}. 

Going forward, let $\mathcal{V}\in\mathbb{R}^{N\times H\times W}$ represent a single-channel thermal video with $N$ thermal frames and a spatial resolution of $H\times W$. $I(n)$ denote the thermal frame at index $n=0,1,...,N-1$. We define a video clip $V(n)$ as the preceding $F$ frames:
\begin{equation}
    V(n) = \{I(n),I(n-1),...,I(n-F+1)\}
\end{equation}

\subsection{Video preprocessing}
\label{subsec:vid_pp}

\begin{figure}[t]
\centering
\begin{minipage}[b]{\linewidth}
  \centering
  \centerline{\includegraphics[width=\linewidth]{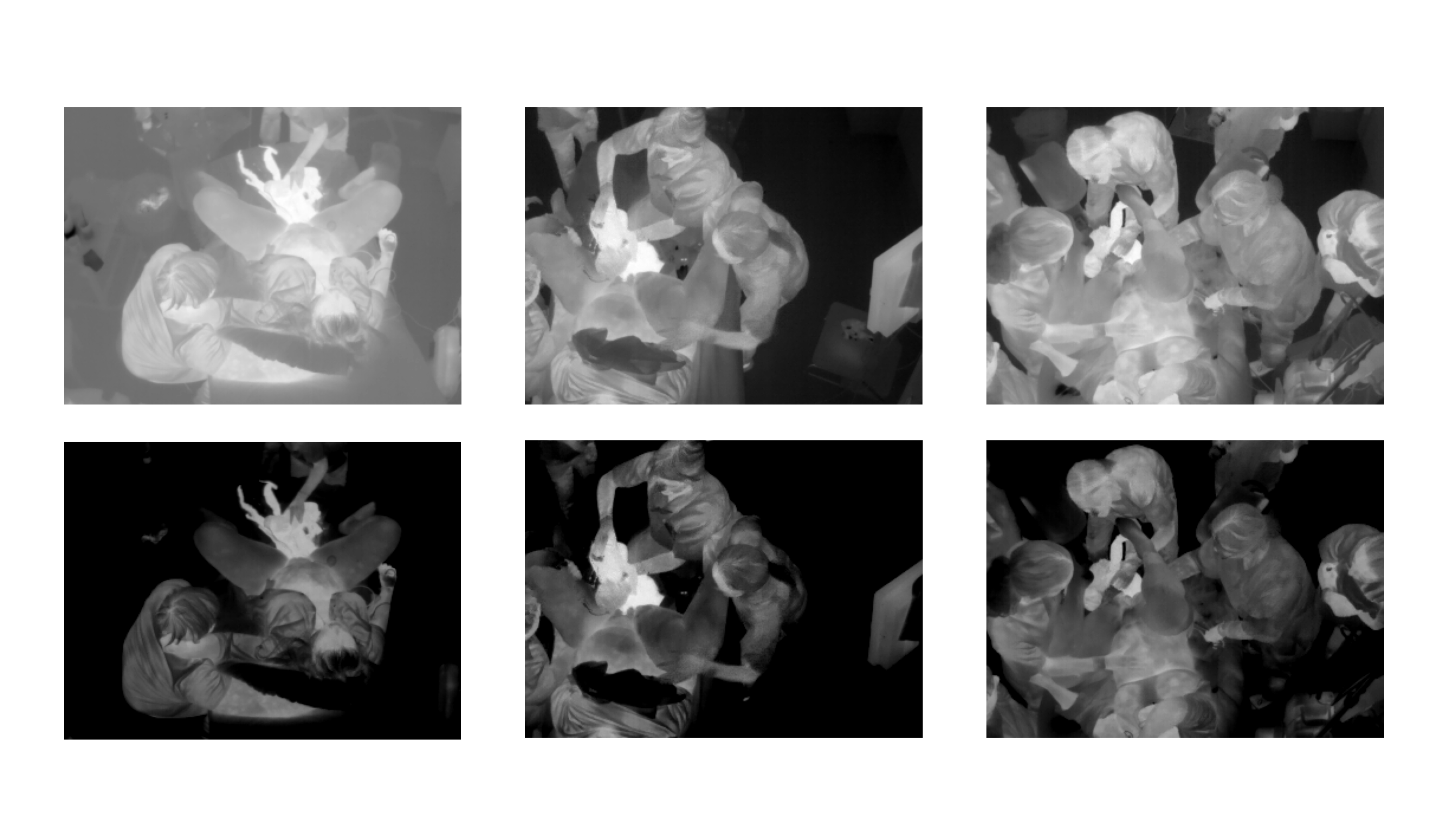}}
\end{minipage}
\caption{Illustration of thermal frames with Max-Min normalization (first row) and our proposed GMM normalization (second row).}
\label{fig:gmm_ex}
\end{figure}

In García-Torres et al. \cite{garcia2022towards}, we explored the challenges of using thermal sensors for ToB detection, finding limitations in relying on absolute temperature values due to several distorting factors such as autocalibration and room temperature. This potential reliability issue required the use of relative temperatures and introduced a normalization challenge for our AI system. To address this, we later proposed a GMM-based normalization method \cite{garcíatorres2024advancingnewborncareprecise} to effectively identify a relevant temperature range based on human skin temperature in each thermal video, preserving the physical meaning of relative temperatures and ensuring more consistency across videos (see Figure~\ref{fig:gmm_ex}).

GMM normalization involves modeling the data distribution of a thermal video using three Gaussian components, each representing three different regions: low temperatures (background), mid temperatures (clothes, hair, bed sheets), and high temperatures (human skin). For each thermal video $\mathcal{V}$, intensity values are extracted every 30 seconds, converted to temperature values, and organized into a one-dimensional vector $\mathbf{v}$, which is used to fit the GMM. The highest mean value $\hat{\mu}_\mathbf{v}$ among the Gaussian components is selected after applying empirical constraints to define a temperature range of interest based on observations from the delivery room. We define the GMM normalization function as $\phi_{\hat{\mu}_\mathbf{v}}$, where $\Tilde{\mathcal{V}} = \phi_{\hat{\mu}_\mathbf{v}}(\mathcal{V})$ represents the process of clipping the temperature values within the defined range of interest and rescaling them to fall between 0 and 1.

To generate video clips, we sample from $\Tilde{V}(n)$ at specific time intervals, mapping the frame index $n$ to the timestamp $t$ according to the frame rate $f_r$ and a constant temporal stride $\tau$. The resulting sampled clip set $x(t)\in\mathbb{R}^{F\times H\times W}$ is defined as:
\begin{equation}
    x(t) = \Tilde{V}(\lfloor f_r \cdot t \rfloor) \quad t=t_0,t_0+\tau,t_0+2\tau,...,\lfloor N/f_r \rfloor
\end{equation}
where $t_0=\lfloor F / f_r \rfloor$ is the first valid timestamp and $\lfloor \cdot \rfloor$ denotes the floor function. Values of $t$ lower than $t_0$ are not considered due to boundary problem. 

\subsection{CNN backbone}
\label{subsec:backbones}

We selected I3D \cite{carreira2017quo}, X3D \cite{feichtenhofer2020x3d}, and MoViNet \cite{kondratyuk2021movinets} as feature extractors for activity recognition due to their effectiveness in spatiotemporal learning and video analysis. I3D's inflated 3D convolutions enable it to capture rich spatial and temporal features, which is crucial for accurately recognizing complex activities. X3D provides a scalable and efficient architecture, making it ideal for balancing performance and computational cost. MoViNet (stream) offers an efficient, real-time solution, suitable for low-latency environments. CNN-based models were chosen over transformers due to the small size of our dataset and the limited computational resources available in clinical settings, where real-time performance is essential.

For each backbone, we adjust its input size to match our thermal video frame resolution. We also adapt the original architecture's classifier layer for the binary classification task. For using pre-trained weights, our single-channel input data is converted into RGB data by expanding the intensity values to the three channels. 

\subsection{Inference process}
\label{subsec:inference}

To perform inference, a full thermal video is streamed into the model as $x(t)$ using a sliding window with a constant temporal stride. We represent the positive prediction score for birth detection as $\hat{y}$ for simplicity. Given a model $\varphi$ with parameters $\theta$, the temporal probability score is defined as $\hat{y}(t) = \varphi_\theta(x(t))$.

This temporal score is utilized for birth detection. To mitigate possible noise and risk of false detections, we filter this score signal with a smoothing FIR filter of size $K$ with filter coefficients $h(k)=\frac{1}{K}$, $\forall k$. The filtered score signal is computed as follows:
\begin{equation}
    \hat{y}_h(t) = \sum_{k=0}^{K-1}h(k)\cdot\hat{y}(t-k)
\end{equation}
To estimate the ToB, we identify the first timestamp where the filtered scores exceed a predetermined confidence threshold $\gamma$.
\begin{gather}
    \hat{T}_{birth}=min\{t\,|\,\hat{y}_h(t)\geq\gamma\}
\end{gather}

\section{Experiments}
\label{sec:exp}

In this study, we conduct two distinct experiments. First, we implement a modeling pipeline to train, validate, and test our model on thermal video clips for a binary classification task: ToB or No Birth (NB). In the second experiment, we select the model with the best performance from the first experiment to run inference on the test set and estimate ToB. The test set consists of 25 manually selected thermal videos, keeping the same distribution of the maternal position during birth as the dataset. Videos involving twins have been excluded from the test set. The remaining videos are allocated to the training and validation sets, following an 85\%/15\% split.

\subsection{Exp. 1. Video clip classification}
\label{subsec:clip_exp}

\subsubsection{Modeling data}
\label{subsubsec:data}

During the model development, thermal videos are trimmed into short clips, with labels assigned based on the manual annotations. Clips are sampled using a window size of size of 4.5 seconds ($F=37$ frames), wide enough to capture the birth dynamics but allowing for a fine-grained estimation of the time of birth. We set the stride to $\tau=0.5$ seconds to generate more ToB samples, as birth occurs over a very short time span. To guarantee birth dynamics are captured effectively, we ensure that the birth event is not positioned near the window boundaries.

To address the inherent class imbalance in the dataset, we downsample the NB class by extracting a video clip every 30 seconds from continuous sequences of NB. This strategy helps retain NB variability while significantly reducing the number of instances, though full data balance is not achieved, making the use of a weighted cross-entropy loss function necessary during training.

\subsubsection{Implementation details}
\label{subsubsec:taining}

Binary cross-entropy \cite{bishop2006pattern} is employed as the loss function. To mitigate the impact of class imbalance, we estimate the inverted class weight $w_c$ for each class $c\in\{0 \text{ (NB)},1 \text{ (ToB)}\}$. Representing the ground truth of the data sample index $q$ as $y_q$, the weighted binary cross-entropy loss function $\mathcal{L}$ is defined as:
\begin{align}
    \mathcal{L}(y_q,\hat{y_q}) = w_1 y_q \log(\hat{y}_q) + w_0 (1 - y_q) \log(1 - \hat{y}_q)
\end{align}
We then deploy and evaluate several binary video-based models to classify ToB and NB video clips using different CNN backbones as feature extractors. For each backbone, we train for a maximum of 100 epochs with early stopping. Adaptive Moment Estimation (Adam) with $\beta_1$=0.9 and $\beta_2$=0.999 is utilized. Weight decay of 0.97 is applied every 1k steps with a learning rate of 1e-5. We also use a moving average with a 0.9999 decay rate. Three Teslas V100 GPUs with 32 GB RAM are employed, setting a batch size of 16 per GPU. 

Data augmentation is performed at video clip level during training. Brightness and contrast augmentation are applied to modify the overall luminance and the intensity differences. Additionally, we apply random left-right flipping. We also implement temporal cropping by randomly picking 25 consecutive frames in the clip. During validation and test, we pick the 25 frames centered in the video clip. 

We assess precision and recall as evaluation metrics for binary classification \cite{lipton2014optimal}. We also employ Matthew's Correlation Coefficient (MCC) \cite{chicco2020advantages} as a more comprehensive metric

\subsection{Exp. 2. Time of birth estimation}
\label{subsec:tob_exp}

During inference, sampled clips $x(t)$ are generated by applying a sliding window of the previous $F=25$ frames (3 seconds) to meet the model's input requirements. As second precision is seen as precise enough for a ToB detector, we use a stride of $\tau=1$ second to improve temporal precision. We apply a FIR filter of size $K=3$ samples to the prediction scores $y(t)$, resulting in the filtered scores $\hat{y}_h(t)$, followed by a confidence threshold analysis to estimate ToB.

For evaluation purposes, we define the error $err$ as the time difference between $\hat{T}_{birth}$ and the manual annotated ToB ($T_{birth}$):
\begin{equation}
    err = \hat{T}_{birth} - T_{birth}
\end{equation}
A positive $err$ indicates that the predicted ToB occurs after the actual birth, while a negative $err$ implies that the predicted ToB is before the birth. Additionally, we use the absolute error $|err|$ to compute statistical measures, including the first quartile (Q1), median (Q2), third quartile (Q3), and mean value.

\section{Results \& Discussion}
\label{sec:R&R}

Table~\ref{tab:exp} summarizes the evaluation results for video clip classification. In this experiment, all the models demonstrate strong recall despite the class imbalance problem, effectively identifying birth events. However, our primary focus is on minimizing the FP instances to obtain a more precise ToB estimate. In this respect, MoViNet-A2 outperforms other 3D CNN backbones, achieving higher precision and superior overall performance, as reflected by the MCC score. It is worth noticing the more efficient architecture of MoViNet and X3D compared to I3D, achieving competitive performance with significantly fewer parameters.

\begin{table}[t]
\centering
\small
\begin{tabular}{lcccccc}
\hline
\multicolumn{1}{c}{Backbone} & Params (M) & Precision & Recall & MCC \\ \hline \hline
I3D & 12.3 & 0.864 & 0.921 & 0.887 \\
X3D-M & 2.9 & 0.759 & 0.829 & 0.783 \\
MoViNet-A0 & 2.5 & 0.806 & \textbf{0.987} & 0.886 \\
MoViNet-A1 & 5.2 & 0.833 & 0.855 & 0.836 \\
\textbf{MoViNet-A2} & 6.1 & \textbf{0.914} & 0.974 & \textbf{0.94} \\ \hline
\end{tabular}
\caption{Classification performance using different 3D CNN backbones in Exp. 1. ToB is used as the positive class. The number of parameters includes the backbone and the classifier (in millions).}
\label{tab:exp}
\end{table}

\begin{figure}[t]
\centering
\begin{minipage}[b]{\linewidth}
  \centering
  \centerline{\includegraphics[width=.95\linewidth]{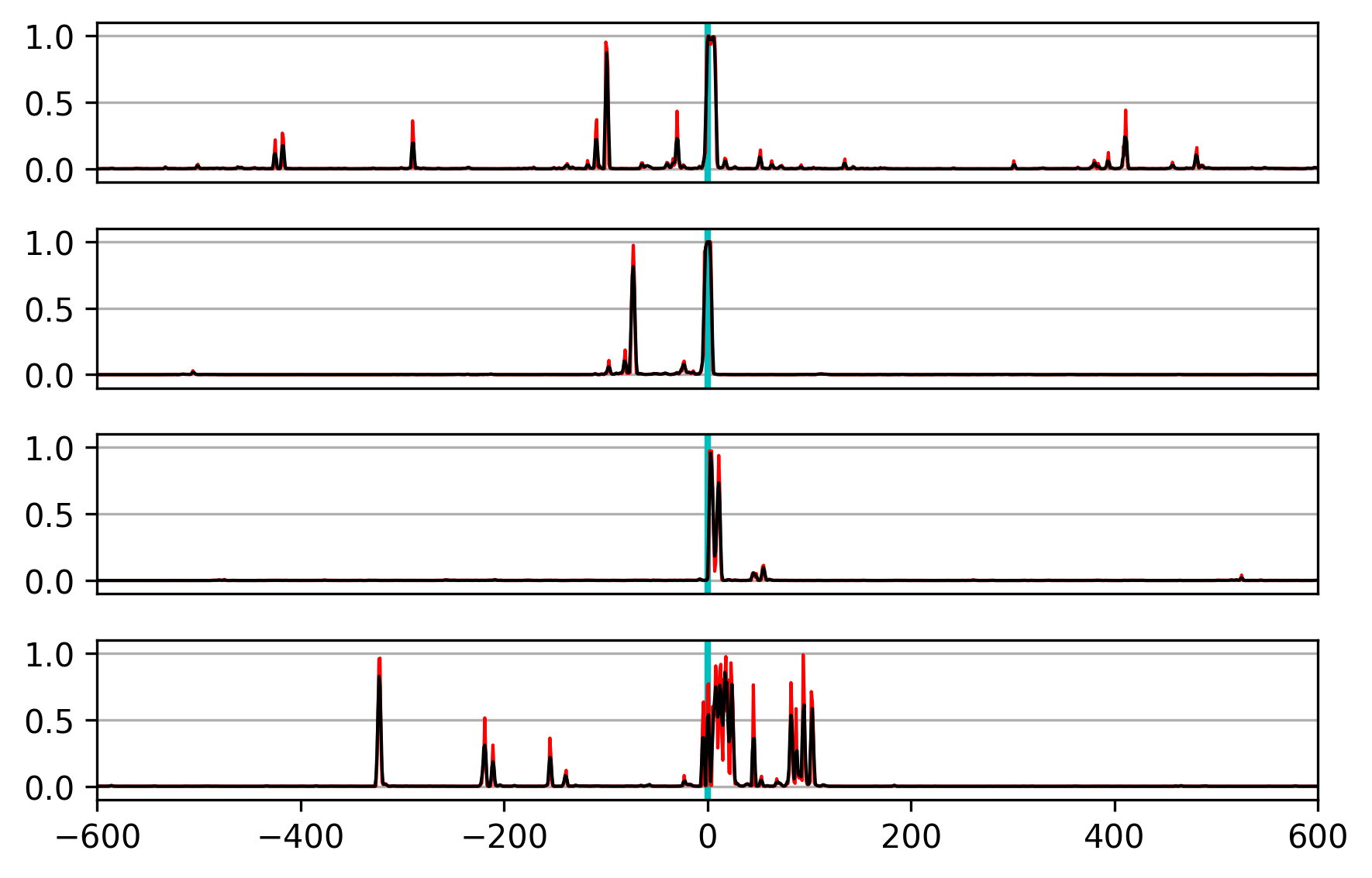}}
\end{minipage}
\caption{Visualization of the ToB probability score in four thermal videos in the range $\pm$10 minutes centered around the manual annotated ToB (in blue). In red, the raw score generated by our MoViNet-A2. In black, the postprocessed score using a FIR filter.}
\label{fig:comb_score}
\end{figure}

\begin{figure}[t]
\centering
\begin{minipage}[b]{.48\linewidth}
  \centering
  \centerline{\includegraphics[width=\linewidth]{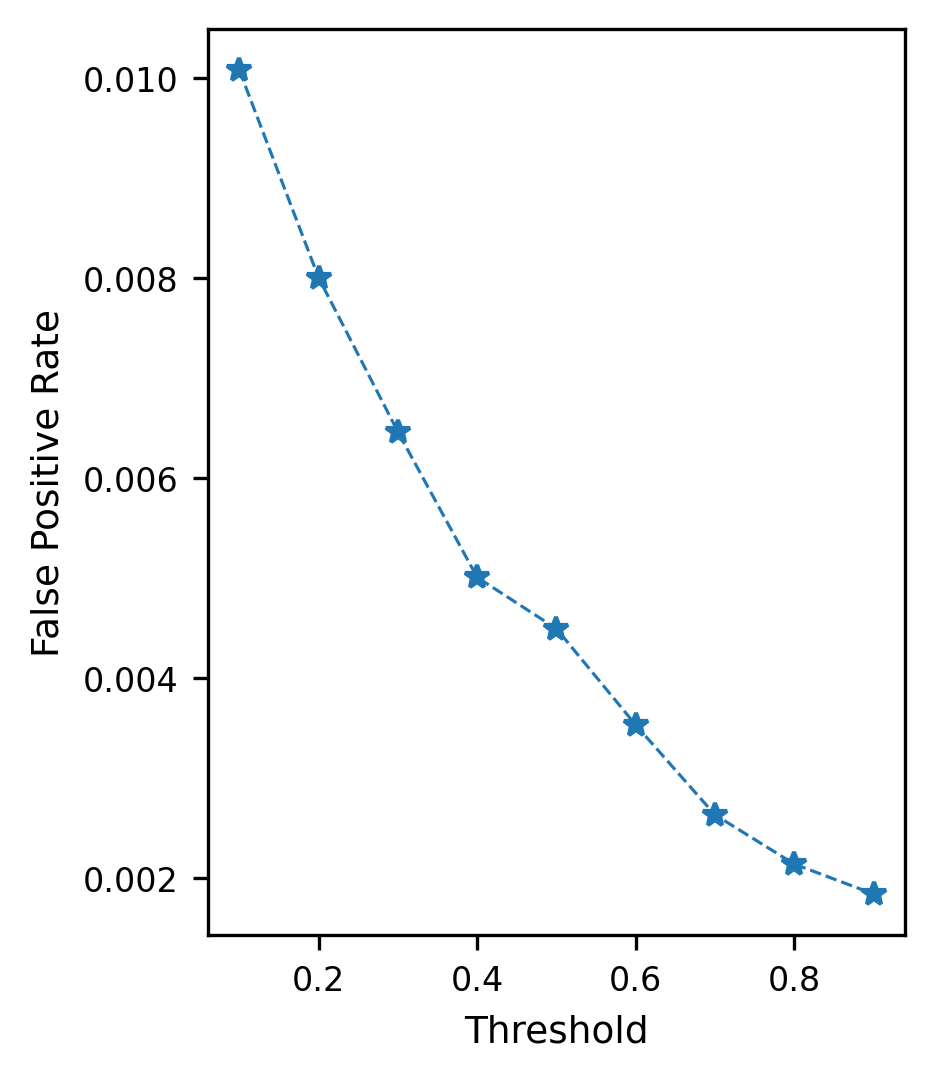}}
  \subcaption{FPR}
  \label{fig:fpr}
\end{minipage}
\begin{minipage}[b]{.44\linewidth}
  \centering
  \centerline{\includegraphics[width=\linewidth]{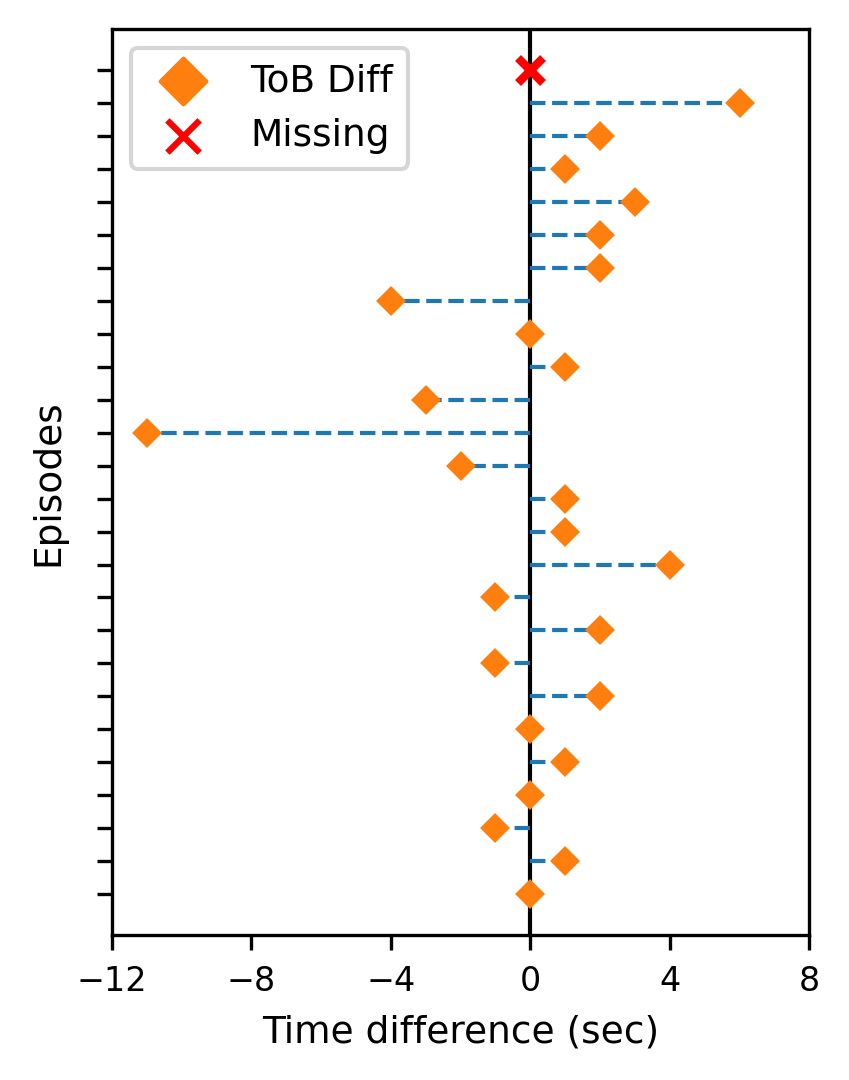}}
  \subcaption{ToB difference}
  \label{fig:tob_diff}
\end{minipage}
\caption{(a) Evaluation of FPR at various confidence thresholds using MoViNet-A2. For higher thresholds, our model exhibits superior capabilities in identifying the ToB. (b) Time difference ($err$) between the predicted and the annotated ToB for the test set videos. If the predicted ToB is not found by our system, it is marked as ``Missing''.}
\label{fig:fpr-tob_diff}
\end{figure}

\begin{table}[t]
\centering
\small
\begin{tabular}{lcccccc}
\hline
\multicolumn{1}{c}{Method} & FIR & Q1 & Q2 & Q3 & Mean & B.F.\\ \hline \hline
Image-based\cite{garcíatorres2024advancingnewborncareprecise} & \ding{51} & 0.8 & 3.2 & 13.7 & 87.8 & 100\% \\
Video-based & \ding{55} & 1 & 1 & 5 & 45.8 & 100\% \\
Video-based & \ding{51} & 1 & 1.5 & 2.25 & 2.1 & 96\% \\ \hline \hline
\end{tabular}
\caption{Statistical performance on the test set in Exp. 2. All values are provided in seconds. Q values represent the quartiles. B.F. indicates the percentage of test videos where $\hat{T}_{birth}$ was found. The drop in B.F. corresponds to the "Missing" value in Figure~\ref{fig:tob_diff}.}
\label{tab:boxplot}
\end{table}

In Exp. 2, we evaluate the capability of our system to estimate the ToB. As detailed in Section~\ref{subsec:inference}, we first filter the prediction scores, followed by an analysis of the confidence threshold. Examples of the predicted and filtered scores for this model are presented in Figure~\ref{fig:comb_score}. Figure~\ref{fig:fpr} depicts the False Positive Rate (FPR) of $\hat{y}_h(t)$ at various thresholds, indicating that higher thresholds lead to a significant reduction in false birth detections.

By setting a threshold of $\gamma=0.9$, we examine the time difference between our predicted ToB and the manual annotations across the test set, as illustrated in Figure~\ref{fig:tob_diff}. Table~\ref{tab:boxplot} presents a comparison of the statistical results from the same 25 test videos using our video-based approach and the image-based method \cite{garcíatorres2024advancingnewborncareprecise}, both evaluated with $\gamma=0.9$, showing a notable improvement of the video-based approach. By capturing temporal birth dynamics, our new system effectively addresses issues associated with relying solely on spatial information such as the presence of static warm elements or limited visibility of the newborn, resulting in a more fine-grained and precise ToB estimate. It is important to note that the dataset used in this work is larger than the one used in the image-based method, as the project is ongoing and more data has been recorded.

We also evaluate the effect of filtering the prediction scores for ToB estimation, which significantly improves the model's performance, achieving results that closely align with the ground truth in 96\% of the test videos. In contrast, combining the FIR filter with a high-confidence threshold shows an increased risk of missing the ToB. The specific missing case in Figure~\ref{fig:tob_diff} occurs due to hands-and-knees maternal position, where the visibility of the birth is often significantly limited. Having a fallback system using the image-based approach could be a potential solution to mitigate this issue.

It is worth mentioning that our GMM normalization is performed on an entire video, being suitable for debriefing and research. For real-time decision support, it should be performed causally with periodically updated parameters. We will leave this for future work.

\section{Conclusion}
\label{sec:conclusion}

In this work, we present a spatiotemporal AI-based approach for estimating the ToB using thermal video. Our system achieves a precision of 91.4\% and a recall of 97.4\% in identifying the birth within thermal video clips and estimates ToB with a median absolute deviation of 1 second from manual annotations. The use of thermal videos protects the privacy of healthcare providers and mothers while enabling accurate ToB detection. This data could also help assess whether breathing aid interventions like stimulation are initiated at the bedside. In future work, we will combine the image-based and video-based approaches to mitigate the risk of missing the birth. We will also culminate the creation of NewbornTimeline -- a system capable of automatically generating detailed NRAA timelines, including ToB detection and AI-based activity recognition from resuscitation videos. 

We believe that the integration of thermal imaging and AI-based activity recognition from video can be extended to other non-diagnostic healthcare applications, such as documenting activities in emergency care or security surveillance, while maintaining privacy for both healthcare providers and patients.

\section{Compliance with ethical standards}
\label{sec:ethics}

This study was performed in line with the principles of the Declaration of Helsinki. The study has been approved by the Regional Ethical Committee, Region West, Norway (REK-Vest), REK number: 222455. The project has been recommended by Sikt - Norwegian Agency for Shared Services in Education and Research, formerly known as NSD, number 816989. Informed consent was obtained from all mothers involved in the study.

\vspace{-0.1cm}
\section{Acknowledgments}
\label{sec:acknowledgments}
\vspace{-0.08cm}

The NewbornTime project is funded by the Norwegian Research Council (NRC), project number 320968. Additional funding has been provided by Helse Vest, Fondation Idella, and Helse Campus, Universitetet i Stavanger. Study registered in ISRCTN Registry, number ISRCTN12236970. 

\vspace{-0.3cm}
\bibliographystyle{IEEEbib}
\bibliography{strings}

\end{document}